\documentclass[sigconf]{acmart}

\AtBeginDocument{%
  \providecommand\BibTeX{{%
    \normalfont B\kern-0.5em{\scshape i\kern-0.25em b}\kern-0.8em\TeX}}}

\copyrightyear{2024}
\acmYear{2024}
\setcopyright{acmlicensed}\acmConference[CODS-COMAD 2024]{7th Joint International Conference on Data Science \& Management of Data (11th ACM IKDD CODS and 29th COMAD)}{January 4--7, 2024}{Bangalore, India}
\acmBooktitle{7th Joint International Conference on Data Science \& Management of Data (11th ACM IKDD CODS and 29th COMAD) (CODS-COMAD 2024), January 4--7, 2024, Bangalore, India}
\acmDOI{10.1145/3632410.3632464}
\acmISBN{979-8-4007-1634-8/24/01}

\usepackage{url}
\usepackage{amsmath}
\usepackage{booktabs}
\usepackage{enumerate}
\usepackage{algorithmic}
\usepackage[ ruled,vlined, norelsize]{algorithm2e}
\usepackage[switch]{lineno}

\usepackage{amsmath}

\usepackage{amsfonts} 
\usepackage{todonotes}
\usepackage{xspace}
\usepackage{multirow}
\newcommand{\ch}{\mathbf{h}\xspace}

\newcommand{\x}{\mathbf{x}\xspace}
\newcommand{\s}{\mathbf{s}\xspace}

\newcommand{\btheta}{\boldsymbol{\theta}\xspace}
\newcommand{\bg}{\mathbf{g}\xspace}

 \newtheorem{defn}{\textbf{Definition}}

\def\tgnns{\textsc{Tgnn}\xspace}
\def\tgat{\textsc{TGAT}\xspace}
\def\tgn{\textsc{TGN}\xspace}
\def\gat{\textsc{GAT}\xspace}
\def\pint{\textsc{PINT}\xspace}

\begin{document}

\title{Robust Training of Temporal GNNs using Nearest Neighbours based Hard Negatives}

\author{Shubham Gupta}
\affiliation{%
  \institution{Indian Institute of Technology, Delhi}
  \city{New Delhi}
  \country{India}
}
\email{shubham.gupta@cse.iitd.ac.in}

\author{Srikanta Bedathur}
\affiliation{%
  \institution{Indian Institute of Technology, Delhi}
 \city{New Delhi}
  \country{India}
}
\email{srikanta@cse.iitd.ac.in}



\renewcommand{\shortauthors}{Gupta, et al.}

\begin{abstract}
Temporal graph neural networks \tgnns have exhibited state-of-art performance in future-link prediction tasks. Training of these TGNNs is enumerated by uniform random sampling based unsupervised loss. During training, in the context of a positive example, the loss is computed over uninformative negatives, which introduces redundancy and sub-optimal performance. In this paper, we propose modified unsupervised learning of \tgnns, by replacing the uniform negative sampling with importance-based negative sampling. We theoretically motivate and define the dynamically computed distribution for a sampling of negative examples. Finally, using empirical evaluations over three real-world datasets, we show that \tgnns trained using loss based on proposed negative sampling provides consistent superior performance.
\end{abstract}




\maketitle

\section{Introduction and Related Works}


Interactions between entities are common in domains such as e-commerce \cite{amazon}, finance \cite{bitcoin}, and online forums(communities) \cite{community}. Precise modeling of the future preferences of entities in such graphs has many applications e.g. recommendations, anomaly detection, clustering, influence maximization, etc. These preferences are dynamic, i.e., influenced by the past interactions of the target entity and its neighborhood's behavior. Thus, learning user biases requires an efficient representation to encode structural and temporal information provided in the interaction graph.

Temporal Graph Neural Networks (\tgnns) \cite{dysat,dyngem,tNodeEmbed,evolvegcn,MDNE,FiTNE,HTNE,tigecmn,dyrep,tgat,tgn,caw,ige,tigger,grafenne,gupta2022survey} have been proven very successful to encode the future biases of entities by effective sharing of trainable filters across temporal neighborhoods of  nodes in a temporal graph. This encoded information, a low dimensional representation of each node, jointly models the structural and temporal characteristics and is effective in downstream tasks such as dynamic node classification and future link prediction. \tgnns are trained using stochastic gradient descent on unsupervised loss computed over past interactions. The following negative sampling-based unsupervised loss \cite{graphsage} is the common choice for training parameters of Graph Neural Networks(\textsc{GNNs}).
\begin{equation}
\mathcal{L}_v = - \log \sigma(\ch_v^T\ch_u) - \mathcal{Q}\mathbb{E}_{v_n \sim P_n} \log \sigma(-\ch_{v_n}^T \ch_v)
\label{eq:negative_sampling}
\end{equation}

where $u$ is a neighbor of node $v$, $\mathcal{Q}$ is a hyper-parameter which denotes the no. of negative nodes to be sampled, and $P_n$ is a noise distribution over nodes assumed to be uniform\footnote{This assumption is specific to \textsc{GNN} and \tgnns methods.} and independent of target node $v$.  $\ch_v$ is the learned representation of node $v$. Eq. \ref{eq:negative_sampling} is modified for \tgnns as follows. 

\begin{multline}  
\mathcal{L}(u,v,t) = - \log \sigma(\ch_v^T(t^-)\ch_u(t^-)) \\ -\mathcal{Q} \mathbb{E}_{v_n \sim P_n} \log \sigma(-\ch_{v_n}^T(t^-) \ch_v(t^-))
\label{eq:negative_sampling_time}
\end{multline}

where loss is computed for every interaction ($u$,$v$) at time $t$. $\ch_v(t^-)$ is a low dimensional representation estimated using node $v$'s interactions before time $t$.  This representation encodes the future preferences of node $v$ at time $t$ and is utilized in downstream tasks such as recommendations and classifications. In practice, \tgnns assumes $\mathcal{Q}=1$ and approximates eq. \ref{eq:negative_sampling_time} as follows.
\begin{multline}
\mathcal{L}(u,v,t) = \mathcal{L}(u,v,v^-,t) = - \log \sigma(\ch_v^T(t)\ch_u(t)) \\ - \log \sigma(-\ch_{v^-}^T(t) \ch_v(t))
\label{eq:negative_sampling_time_approx}
\end{multline}
where $v^- \sim p_{-}$, $p_{-}$ is uniform distribution over nodes.

\tgnns trained using eq. \ref{eq:negative_sampling_time_approx} are sub-optimal for downstream tasks due to the use of uniform random sampling distribution $p_{-}$ yielding the following problems.\\

\begin{figure*}[t]
\centering
\includegraphics[scale=0.38]{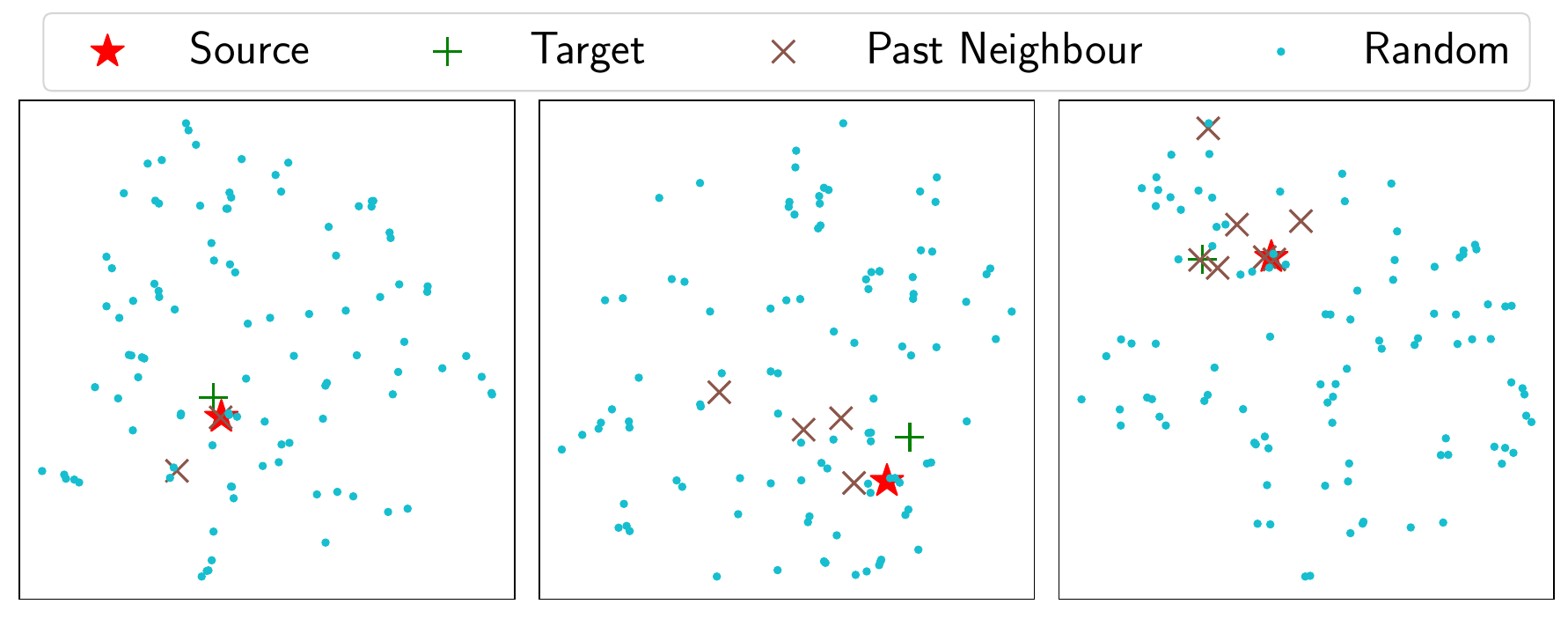}  
\caption{2D T-SNE \protect\cite{tsne} representation of node embeddings for various interactions between source and target node pairs. These representations are computed at time $t$ just before the interactions.  It is clearly seen that learned temporal embeddings for past nodes and target nodes are nearby, and often target node is farther than past nodes from the source nodes. Also, the representation of a few random nodes is closer to the source node than the target node. This results in sub-optimal performance of \tgnns in recommendation tasks.}
\label{fig:tsne_plots}
\end{figure*}

$\bullet$ \textbf{Inadequate differentiation between past and future neighbour nodes:} Nodes in static graphs have multiple neighbors. Thus, embeddings learned using equation \ref{eq:negative_sampling} are optimal since they assign similar embeddings to nearby nodes. On the contrary, in a continuous time interaction graph, every node (source) can interact with only one other node (target) at any given time. This is a more challenging task and is not incorporated into the loss. Thus, learned embeddings don't produce sufficiently different representations of past neighbors and future neighbors of a given source node. We also observe this in figure \ref{fig:tsne_plots}, where we plot the $2$D T-SNE \cite{tsne} representations of all nodes at various timestamps after training the \tgnns \cite{tgn} on the Wikipedia dataset \cite{jodie}. Each timestamp corresponds to an interaction between a sampled source node and its target node. We distinctly see that the learned embeddings reflect past neighbors nearer to the source node instead of the target node, highlighting the stated problem.

$\bullet$ \textbf{Similar temporal neighbors} \tgnns requires node features as input. And as many nodes consist of similar features with similar temporal neighbors, learned embeddings may not differentiate enough between such nodes. This can be visualized in figure \ref{fig:tsne_plots} where many nodes are between source and target nodes. These issues become severe in recommendation tasks where precision in top-ranked results is desirable. 

While we discuss these $2$ as possible problems, we postulate there can be more causes for the sub-optimal embedding distribution in figure \ref{fig:tsne_plots}. Thus, the ideal solution lies in designing the negative sampling distribution $p_{-}$ in a way making no assumption about the biases in the data. 


\subsection{Existing Works}
The selection of negatives during training has not drawn any attention in \tgnns, though it has been extensively studied in \textit{vision} and  \textit{natural language processing}. Specifically, It has been observed that uniformly sampled random negative examples (texts/images) are too far from the source(query/anchor) and target(positive) representations making their contribution negligible in loss computation during training. Therefore, modification of random negative samples within unsupervised learning has been shown in {information retrieval} \cite{karpukhin-etal-2020-dense,ance,rocketqa,rocketqav2,lin-etal-2021-batch,tas,hard_negatives_sigir} and image classification  \cite{chuang2020debiased,ho2020contrastive,Kalantidis2020HardNM,robinson2020contrastive} methods. A whole other body of research is on contrastive learning methods for unsupervised training. These methods avoid the cost of labeling large-scale datasets and typically generate pseudo-labels by augmenting/modifying the source and target examples. This is not the focus of our work, and we direct interested readers to a literature survey \cite{jaiswal2020survey}.

\subsubsection{NLP} The methods in this domain propose replacing random negative examples with hard negatives. Commonly, the negative data point is sampled from a distribution dependent on the query. \cite{karpukhin-etal-2020-dense} uses negative from \textit{TF-IDF} based \textit{BM25} text retriever to train the query and passages dual encoders. In contrast, \cite{ance} computes the closest example from the query using cosine similarity of representations learned during training and uses it as a hard negative. \cite{rocketqa,rocketqav2} observes that hard negatives samples are often correct paragraphs not labeled as positive by annotators. They propose to de-noise the hard negatives using a strong cross-encoder teacher model. 

\subsubsection{Vision} \cite{robinson2020contrastive} proposes using negative samples labeled differently from source images and closest to it in the embedding space. \cite{Kalantidis2020HardNM} designs the negative examples by combining the embeddings of hard negative examples. \cite{chuang2020debiased} proposed a de-biased method for sampling hard-negatives by correcting for the possibility of sampling from the same label as the source example in a set-up where labels of examples are unknown. \cite{ho2020contrastive} instead replaces the hard negatives with adversarial examples learned during the training.  

\subsubsection{Graphs} Hard random negatives are relatively unexplored in graph neural networks, specifically in temporal graphs neural networks. For unsupervised static graph representation learning, \cite{Zhu2021StructureAwareHN} has proposed personalized page-rank \cite{ppr} based hard-negatives by computing page rank scores of each node from the source node and picking the top-k nodes with the highest scores.  \cite{pinsage} develops a {Graph Neural Network} for a large-scale recommendation system and remarks that during training model will easily distinguish negative nodes randomly sampled out of billion nodes from few relevant nodes and uses hard negatives similar to \cite{Zhu2021StructureAwareHN}. \cite{ns_graph} proposes to sample negative node using a distribution sub-linearly correlated with positive node distribution given query node. This is not relevant to \tgnns as the source and query node are fixed.  More recent works \cite{graphdc,cuco,progcl,Xu2021SelfsupervisedGR,grace,dgi} have proposed contrasting learning methods for static graphs. These contrastive methods create multiple-augmented views of nodes and minimize a contrastive loss to train GNNs that utilize such views of positive and negative samples. Recent papers \cite{milocalglobal,GMI} further extends this to maximize the mutual information (MI) between global and local graph information. Our work doesn't relate to these settings as we work with a single view of each node and especially analyze the impact of hard negatives in the original loss function of \tgnns.

 
While the use of hard negative in \tgnns is unexplored, applying negative mining methods of existing works in temporal settings is not straightforward, e.g., a node that is not a hard negative for a source-target node pair at time $t$ may become one at $t'>t$. This needs an assumptions-free design of negative node distribution to train \tgnns.

\subsection{Contribution}
To the best of our knowledge, this is the first work to research hard negative nodes for training \tgnns. We summarize our contribution as follows:
\begin{itemize}
    \item We theoretically analyze the connection between hard negative samples during training \tgnns with parameter convergence and propose an assumption-free and domain-agnostic novel negative node sampling distribution.
    \item We evaluate the proposed distribution on $3$ real-world temporal interaction datasets by integrating it with \textsc{TGAT} \cite{tgat}, \textsc{TGN} \cite{tgn} and establish significant improvement in recommendation task setup. 
\end{itemize}


\section{Background: Temporal graph neural networks}
This section introduces the preliminary ideas behind Temporal Graph Neural Networks (\tgnns). We first define a temporal interaction graph $\mathcal{G}$, then explain the temporal neighborhoods integral to feature propagation and aggregation in \tgnns. Finally, we define the message-passing equations to compute the temporal node representations. 
\begin{defn}[Continuous time temporal graph]
\label{def:cttg}
A continuous time temporal graph is defined as a stream of interactions $\mathcal{G}=\{e_1,e_2 \ldots \}$ where each interaction  $e_i=(u,v,t,\x_u,\x_{uv},\x_v)$ is a tuple of source node $u \in \mathcal{V}$, target node $v \in \mathcal{V}$, time of interaction $t \in R^+$, node $u$'s feature vector $\x_u \in \mathcal{R}^D$, node $v$'s feature vector $\x_v \in \mathcal{R}^D$ and edge feature vector $\x_{uv} \in \mathcal{R}^D$\footnote{Feature dimension of nodes and edges is assumed to be same for simplicity. The proposed method can easily be modified for different dimensions also}. $\mathcal{V}$ is a set of $N={\mid \mathcal{V}\mid}$ nodes.  For simplicity, we assume undirected interactions.
\end{defn}
The objective of \tgnns is to effectively encode the past information of each node to predict its future behavior. To this extent, we define the temporal neighborhoods representing this past information.
\begin{defn}[Temporal Neighborhood]
\label{def:tng}
The temporal neighborhood of node v at time t is a set of interactions by node $v$ before time $t$. Mathematically,
\begin{equation}
    \mathcal{N}_v(t) = \{e=(u,v,t',\x_u,\x_{uv},\x_v) \in \mathcal{G} \mid t' < t\}
\end{equation}
\end{defn}
\textbf{Temporal graph messaging passing and aggregation}: \tgnns compute the node representation of node $v$ at time $t$ by applying consecutive $L$ layer of graph neural network on its $L$ hop \textit{temporal} neighborhood. Specifically,  $0^{th}$ layer embeddings of node $v$ and its $L$ hop temporal neighbors are initialized using their node features.
\begin{equation}
    \ch_u^0(t) = \x_u(t)+\s_u(t^-) \;\;\forall u \in v \cup \mathcal{N}_v(t)
\end{equation}
where $\x_u(t)$ is latest features vector of node $u$ till time $t$, $\s_u(t)$ is an optional memory vector introduced by \tgn \cite{tgn}, which stores the compressed history of node $u$ before time $t$. This vector is computed using sequential neural nets such as \textsc{LSTM}\cite{lstm}, or \textsc{GRU}\cite{gru}. To compute the $l^{th}$ layer representation of a given node $v$, \tgnns compute messages from its temporal neighbors and aggregate them as follows:
\begin{align}
\nonumber \psi^l(e=&(u,v,t',\x_u,\x_v,\x_{uv}),t) \\&= \psi^l(\ch_u^{l-1}(t),\ch_{v}^{l-1}(t),\phi(t-t'),\x_u,\x_v,\x_{uv})
\end{align}
\begin{equation}
\ch_v^l(t) = \sigma^l(\gamma^l\left(\ch^{l-1}_v(t), \{\psi^l(e,t) \;\mid\; \forall e \in \mathcal{N}_v(t)\}\right))
\end{equation}
where $\gamma^l$ is an aggregator of messages from temporal neighbors at $l^{th}$ layer, and $\psi^l$ is a message computation function. Both $\gamma^l$ and $\psi^l$ are deep neural network-based functions. $\sigma^l$ is a non-linear activation function.  Specifically, \tgat \cite{tgat} uses attention-based message aggregation where the query is source node $v$ and key and value are messages from temporal neighbors. \tgn \cite{tgn} uses summation based aggregation.  $\phi(t)$ is time encoding, transforming a real no. to a $d$ dimensional vector. \tgat \cite{tgat} uses {random fourier projections-based} time encoding and \tgn uses \textsc{Time2Vec} \cite{time2vec} based encoding. 

These local temporal operations are applied $L$ times consecutively on $L$ hop neighborhood of node $v$ to output the time-aware representation at time $t$, i.e., $\ch_v(t) = \ch_v^L(t)$. 
This framework is generic, incorporating dynamic node features and edge features. Finally, model parameters $\btheta$ are learned using the unsupervised loss defined in eq. \ref{eq:negative_sampling_time_approx}.
 Given a sequence of $M$ interactions $\mathcal{G}=\{e_1,e_2,e_3 \ldots e_M\}$, eq. \ref{eq:negative_sampling_time_approx} is computed, and gradients are back-propagated over these interactions chronologically.  For more details, we refer to \cite{tgat} and \cite{tgn}.

\section{Methodology}
This section describes the proposed modified training procedure of \tgnns. Specifically, we aim to replace random negative samples with more informative negative samples in eq. \ref{eq:negative_sampling_time_approx} leading to better convergence of model training. To motivate this, we first examine the impact of negative samples on parameter gradients during training using loss defined in eq. \ref{eq:negative_sampling_time_approx}. Following a similar analysis as \cite{Katharopoulos2018NotAS}, we study the model convergence for \tgnns and establish their relationship with gradients. 
\subsection{Connection of Random Negative Samples with Gradients}
Denoting the \tgnns' parameters as $\btheta_i$ after $i^{th}$ gradient step, a plain $\textbf{SGD}$ step is written as follows:
\begin{equation}
    \btheta_{i+1} = \btheta_i - \eta\nabla_{\btheta_i}\mathcal{L}(u,v,v^-,t)
    \label{eq: sgd}
\end{equation}
where $\eta$ is learning coefficient for \textbf{SGD} and $\btheta_{i+1}$ are updated parameters after $i+1^{th}$ gradient update over a given interaction $(u,v,t)$. $v^-$ is the negative node sampled from $p_{-}$ for the loss computation. These updates are done sequentially over interactions in the training period in each epoch. Assuming $\btheta^*$ as optimal parameters of \tgnns, we define the expected parameter convergence rate at $(i+1)^{th}$ gradient step given $\btheta_i$ and a training interaction $(v,u,t)$ as follows:
\begin{equation}
    S =  \mathbb{E}_{p_{-}}\left[\Vert\btheta_i-\btheta^*\Vert^2_2  - \Vert \btheta_{i+1}-\btheta^*\Vert^2_2\right]
\end{equation}
where $p_{-}$ is the probability distribution of sampling the negative node for the interaction $(v,u,t)$ to compute loss in \ref{eq:negative_sampling_time_approx}. We denote the gradient $\nabla_{\btheta_i}\mathcal{L}(u,v,v^-,t)$ as $\bg(v^-)$. Please note that we assume $\bg$ as a function over $v^-$ only. We assume $u,v,t$ to be fixed for our discussion. We expand $S$ as follows:
\begin{equation}
\begin{aligned}
    S &= \mathbb{E}_{p_{-}}\left[ (\btheta_i - \btheta^*)^T(\btheta_i - \btheta^*) - (\btheta_{i+1} - \btheta^*)^T(\btheta_{i+1} - \btheta^*)
  \right]\\
  &= \mathbb{E}_{p_{-}}\left[ \btheta_i^T\btheta_i -2\btheta_i^T\btheta^* 
 -\btheta_{i+1}^T\btheta_{i+1} + 2\btheta_{i+1}^T\btheta^*\right]\\
  \end{aligned}
\end{equation}
Substituting  $\btheta_{i+1}$ using \ref{eq: sgd}, we get
\begin{equation}
\begin{aligned}
S &= \mathbb{E}_{p_{-}}\left[ 2\eta(\btheta_i-\btheta^*)^T\bg(v^-) -\eta^2\bg(v^-)^T\bg(v^-)
  \right]\\
  &= 2\eta(\btheta_i-\btheta^*)^T\mathbb{E}_{p_{-}}[\bg(v^-)] -\eta^2\mathbb{E}_{p_{-}}[\bg(v^-)^T\bg(v^-)]
  \label{eq:parameter_convergence}
\end{aligned}
\end{equation}
The first term is gradient descent speed corresponding to $\btheta_i$. Therefore, we focus on the second term, which is the variance of gradients. Consequently, to achieve better convergence, we need to select the distribution ${p_{-}}$ which minimizes the variance of $\bg(v^{-})$, i.e., $\mathbb{E}_{p_{-}}[\Vert\bg_{v^-}\Vert^2_2]$ assuming fixed $\eta$.  
We now provide the intuition for solving this optimization problem using importance sampling theory. We can estimate the expectation of a function $f$ as follows-
\begin{equation*}
        \mathbb{E}_{p(x)}[f(x)] \simeq \frac{1}{S}\sum_{s=1}^{s=S}f(x_s) \;x_s \sim p(x) 
\end{equation*}
where $p(x)$ is a probability distribution over $x$. This expectation can also be approximated using a different proposal sampling distribution $q(x)$ over $x$ such that $q(x)>0$ where $p(x)>0$ as follows:
\begin{equation*}
\begin{aligned}
    \mathbb{E}_{p(x)}[f(x)] &=  \mathbb{E}_{q(x)}\left[\frac{p(x)}{q(x)}f(x)\right] \\& \simeq \frac{1}{S}\sum_{s=1}^{s=S}\frac{p(x_s)}{q(x_s)}f(x_s) \;\; x_s \sim q(x) 
    \end{aligned}
\end{equation*}
where $q(x)$ is chosen to be proportional to $p(x)f(x)$ to minimize the variance of estimation. 
In \tgnns, $f(v)=\bg(v)$, $S=1$ and $p(v)=\frac{1}{N}\forall v \in V$ i.e. uniform distribution over nodes. Thus, $p_{-}$ should be proportional to the gradient norm over all nodes if they are used as a negative sample during loss computation over the target interaction $(u,v,t)$ i.e.
\begin{equation}
    p_{-}(v^{-}) \propto \Vert\bg({v^{-}})\Vert_2 \quad  \forall v^{-} \in \mathcal{V}.
\end{equation}
This is computationally prohibitive as it entails computing the gradient norm for every $v \in \mathcal{V}$ as a negative example in loss eq. \ref{eq:negative_sampling_time_approx} to estimate $p_{-}$, followed by sampling the negative node from this estimated distribution and finally computing the loss.  Since this loss is computed over all training interactions in chronological order, the estimation of gradient norm-based negative node distribution will have to be repeated for all $M$ interactions during each epoch. 

We now seek to estimate an approximation for this distribution. If a loss is nearer to $0$, then gradient norms are close to $0$ \cite{Katharopoulos2018NotAS} and vice versa. This implies that if loss computed using a negative node $v^{-}$ is approximately $0$, then gradient norm $\Vert \bg({v^-}) \Vert_2$ calculated using $v^{-}$ will also be $0$ and drives uninformative gradient updates. This indicates that \textit{loss is a good proxy for the per-negative node gradient norm if the loss is close to 0}.

That implies the negative sampling probability distribution $p_{-}$ for $v^-$ should be close to $0$ when the loss is approx 0. Using this observation, we now quantify the $p_{-}$ and propose the novel training procedure for \tgnns in the next subsection. 


\subsection{Modified Training Procedures for \tgnns}
Theoretical analysis in the previous subsection shows that a negative example $v^{-}$ should be sampled in proportion to its gradient norm for each interaction to optimize the model convergence.  This is a computationally expensive operation, and the limitation is exacerbated since $p_{-}$ is dependent on model parameters which themselves depend on time $t$ in interaction, rendering computation of $p_{-}$ in an offline mode impossible. Following conditions hold true for calculating the negative sampling distribution $p_{-}$ using the gradient norm for a given interaction $(u,v,t)$.
\begin{equation}
    p_{-}(v^-) \propto \Vert\bg(v^{-})\Vert_2 \not\!\perp\!\!\!\perp \mathcal{L}(u,v,v^-,t) \not\!\perp\!\!\!\perp  t  \not\!\perp\!\!\!\perp \btheta
    \label{eq:random_negative_restriction}
\end{equation}
where $\not\!\perp\!\!\!\perp$ is symbol of non-independence (dependent). Hence, $p_{-}$ can not be computed before the commencement of training or even before each epoch of training. To avoid the computation of gradient norm-based distribution, we seek to approximate it with $\mathcal{L}$ where loss is approximately $0$, and now simplify eq. 3 for $\mathcal{L}(u,v,v^-,t)$.


$\mathcal{L}(u,v,v^{-},t)$ for a given negative sample $v^{-}$ in eq. \ref{eq:negative_sampling_time_approx} is computed via the dot product between embeddings computed at time $t$ of nodes $u$, $v$  and between nodes $u$ and $v^{-}$. We can assume $u$, $v$ and $t$ as constant in order to learn $p_{-}$ for interaction $(u,v,t)$. Hence, eq. \ref{eq:negative_sampling_time_approx} can be re-written as follows:
\begin{equation}
    \mathcal{L}(v^{-}{ \mid }u,v,t) = \text{C}  - \log \sigma(-\ch_{v^-}^T(t) \ch_u(t)) \propto \ch_{v^-}^T(t) \ch_u(t)
\label{eq:score_negative_sample}
\end{equation}
where $\text{C}$ is a constant. If dot product between $\ch_{v^-}(t)$ and $\ch_u(t)$ is high then $\log$$\sigma$$($$-$ $\ch_{v^-}^T$$(t)$ $\ch_u(t))$ is low, making overall $\mathcal{L}(v^{-}{ \mid } u,v,t)$ high as both $\log$ and $\text{sigmoid}(\sigma)$ are  monotonically strictly increasing functions. Thus instead of recomputing eq. \ref{eq:negative_sampling_time_approx} for every $v^{-} \in \mathcal{V}$ to sample a negative example $v^{-}$ for loss computation, we can approximate using eq. \ref{eq:score_negative_sample} by calculating only dot products or cosine similarities between embeddings of node $u$ and all nodes $v^{-} \in \mathcal{V}$. This approximation is applicable even if the loss is computed using \textit{multi-layer-perceptron} followed by a concatenation of node embeddings instead of dot products.  In this case, $ \mathcal{L}(v^{-}{ \mid }u,v,t) \propto \text{MLP}(\ch_{v^-} \mid\mid \ch_u(t))$. The approximation in eq.\ref{eq:score_negative_sample} still requires re-computing the node embeddings for all $v \in \mathcal{V}$ for all interactions (gradient updates) during each training epoch.  Since node embeddings don't change in a short span of time, thus we can recompute embeddings for all nodes only at a pre-defined frequency in each epoch. Next, we formally define the construction of $p_{-}$.

\noindent \textbf{Probability distribution for sampling random negatives for a given interaction} \boldmath$(u,v,t)$\unboldmath: Since $\mathcal{L}(v^{-}{\mid} u,v,t)$ is a good proxy for $p_{-}(v^{-})$ when it reaches close $0$, this implies that negatives nodes which results in low loss values should not be sampled. This directs to the following constraint over a negative node sampling distribution for a given interaction $(u,v,t)$.
\begin{equation}
    p_{-}(v^{-}) \simeq 0 \;\; \forall\; v^- \in V \wedge \mathcal{L}(v^{-}{ \mid }u,v,t) \simeq 0
\end{equation}


Moreover, few negative nodes exist that will result in a positive loss for a given interaction. This can be seen in Figure 1 too. So, we now define a hyper-parameter top-$K$ which describes the no. of most hard negative samples as per equation \ref{eq:score_negative_sample}. Considering that there are no guarantees of negative examples with the highest loss resulting in best parameter convergence, we propose to sample uniformly from these top-$K$ nodes.

To summarize, original \tgnns select negative samples uniformly from node space $\mathcal{V}$ to evaluate eq. \ref{eq:negative_sampling_time_approx}, while our proposed training procedure samples negative node uniformly from top-$K$ nearest nodes of source node $u$.

Formally, we define probability distribution for sampling negative node $v^-$ for an interaction $(u,v,t)$ as follows:
\begin{equation}
    v^ {-} \sim \text{Uniform}( \text{NearestNbrs}^{\text{top-}K}(u) - \{v\})
    \label{eq:final_sampling_dst}
\end{equation}
i.e., we first compute the top-K nearest nodes with the highest dot-product/cosine/MLP score with node $u$ from embeddings of node-set $\mathcal{V}$ at time $t$. We remove node $v$ from the top-K nodes as the negative node can't be the same as the target node and sample uniformly from the remaining set. This is still a $O(N)$ operation but permits \textit{parallel execution in GPUs}. The overall time complexity of the proposed modified training procedure is $O(ENMC)$ compared to the original training procedure time complexity $O(EMC)$ where $E$ is $\#$ of training epochs, $C$ includes all other parameters like embedding size, $\#$ of layers, avg. neighborhood size. $C$ is consistent across both traditional training and proposed training of \tgnns. The additional $N$ in the proposed modified training is due to the negative sampling distribution estimation. 

We note that the major advantage of utilizing uniform sampling from ${\text{top-}K}(u)$ nearest nodes of source node $u$ at time $t$ allows a  potential speed-up during training. Since ${\text{top-}K}(u)$ nodes mostly remain consistent after a few warmup epochs, we can store these ${\text{top-}K}(u)$ for every interaction $(u,v,t)$ and re-use them in consecutive epochs and recompute only at a fixed frequency $F$. So, the computation time complexity of the proposed training can be reduced to $O(EMCN/F)$  from $O(ENMC)$.

Most importantly, \textit{the increased time complexity is applicable only during the training phase, and time complexity remains the same as the original \tgnns during testing}. We also summarize the modified training procedure of \tgnns in algorithm \ref{alg:modified_training}.




\begin{algorithm}[t]
\DontPrintSemicolon
\SetNoFillComment
\scriptsize
\SetKwInOut{Input}{Input}
\SetKwInOut{Output}{Output}
\Input{A continuous temporal interaction graph $\mathcal{G}$ having $M$ interactions $\{e_1,e_2 \ldots e_M\}$ for training, Node embeddings refresh period $P$, Nearest neighbors re-computation frequency $F$, $\#$ of nearest nodes $\text{top-K}$ }
\Output{Trained model parameters $\btheta^{*}$ on $\mathcal{G}$}
 $\btheta \leftarrow \btheta^0$ \tcp*{Initializing the model parameters}
 $epoch\leftarrow 0$\;
 $nn\_dict \leftarrow \{\}$;
 
\Repeat(){stopping criteria}{
    $batch\_id \leftarrow 0$\;
    \ForEach(){$\text{batch}(u,v,t) \in \mathcal{G}$}{

        \uIf{$batch\_id \;\%\;P = 0 \;\;\&\;\; epoch \;\%\;F = 0 $}{
        \tcc{Re-compute node embeddings at time $t$ }
        NodeEmbeddings $\leftarrow$ $\{\text{TGN}_{\btheta}(v,t) \mid \forall v \in \mathcal{V}\}$\;
        }
        $batch\_id \leftarrow batch\_id + 1$\;
        \uIf{$epoch \;\%\;F = 0 $}{
        \tcc{Re-compute $\text{top-K}$ nodes for interaction \(u,v,t\) using NodeEmbeddings }
        $nn\_dict[(u,v,t)] \leftarrow \text{NearestNodes}^{\text{top-K}}(u) - \{u,v\}$ \; 
        }
        
        \tcc{Sampling negative random node uniformly from top-K most-similar nodes of node $u$}
        $v^-$ $\sim$ $nn\_dict[(u,v,t)]$\; 
        $\ch_u^L(t)$, $\ch_v^L(t)$, $\ch^L_{v^{-}}(t)$ $\leftarrow$ $\text{TGN}_{\btheta}(u,t)$,$\text{TGN}_{\btheta}(v,t)$,$\text{TGN}_{\btheta}(v^{-},t)$\; 
        $\mathcal{L}$ $\leftarrow$ $\text{loss}(\ch_u^L(t), \ch_v^L(t), \ch^L_{v^{-}}(t), t)$ \tcp*{Compute loss using eq. \ref{eq:negative_sampling_time_approx}}
        $\btheta \leftarrow \btheta-\alpha \nabla_{\btheta}\mathcal{L}$   \tcp*{SGD step}
}
$epoch \leftarrow epoch+1$\;
  }
  \Return $\btheta$

\caption{Training \tgnns on stream of interactions $\mathcal{G}=\{e_1,e_2 \ldots e_M \}$}
\label{alg:modified_training}
\end{algorithm}

\section{Experiments}
In this section, we examine the proposed method's effectiveness and answer the following questions.

\noindent \textbf{Robust training:} We examine the effectiveness of the proposed hard negatives based \tgnns training by comparing it against the standard \tgnns training. We conduct the experiments over the benchmark of the attributed temporal interaction graphs and empirically show that the proposed training procedure clearly results in better model parameters resulting in superior performance on test data. 


\noindent \textbf{Evaluation against heuristics-based hard negative mining techniques:}  We define two intuitive baselines for hard negative mining. We train \tgnns using negatives selected from these baselines. We show that sampling nodes from such heuristics often perform poorly against the proposed method and even against uniformly random-sampling-based negative examples of standard training.

\noindent \textbf{Sampling uniformly from top-$k$ hard negatives vs. always choosing most hard negative example:} We also answer the question raised while defining the sampling distribution in eq.\ref{eq:final_sampling_dst} that is, can we pick the negative node with the highest loss instead. We show that such a strategy results in sub-optimal performance of trained \tgnns.

Additionally, we analyze the impact of hyper-parameters, such as $\text{top-K}$, periodicity ($P$) of dynamic embedding refresh during each epoch, frequency $F$ of computing $\text{top-K}$ nearest nodes for every training interaction and learning rate $\eta$ as it shows up in \ref{eq:parameter_convergence}. 
Our codebase is available at \url{https://github.com/data-iitd/robust-tgnn}

In the next subsections, we provide dataset details, baseline \tgnns, and performance metrics for evaluation and discuss the results of experiments.
\begin{figure}[b!]
\centering
\includegraphics[width=.4\textwidth]{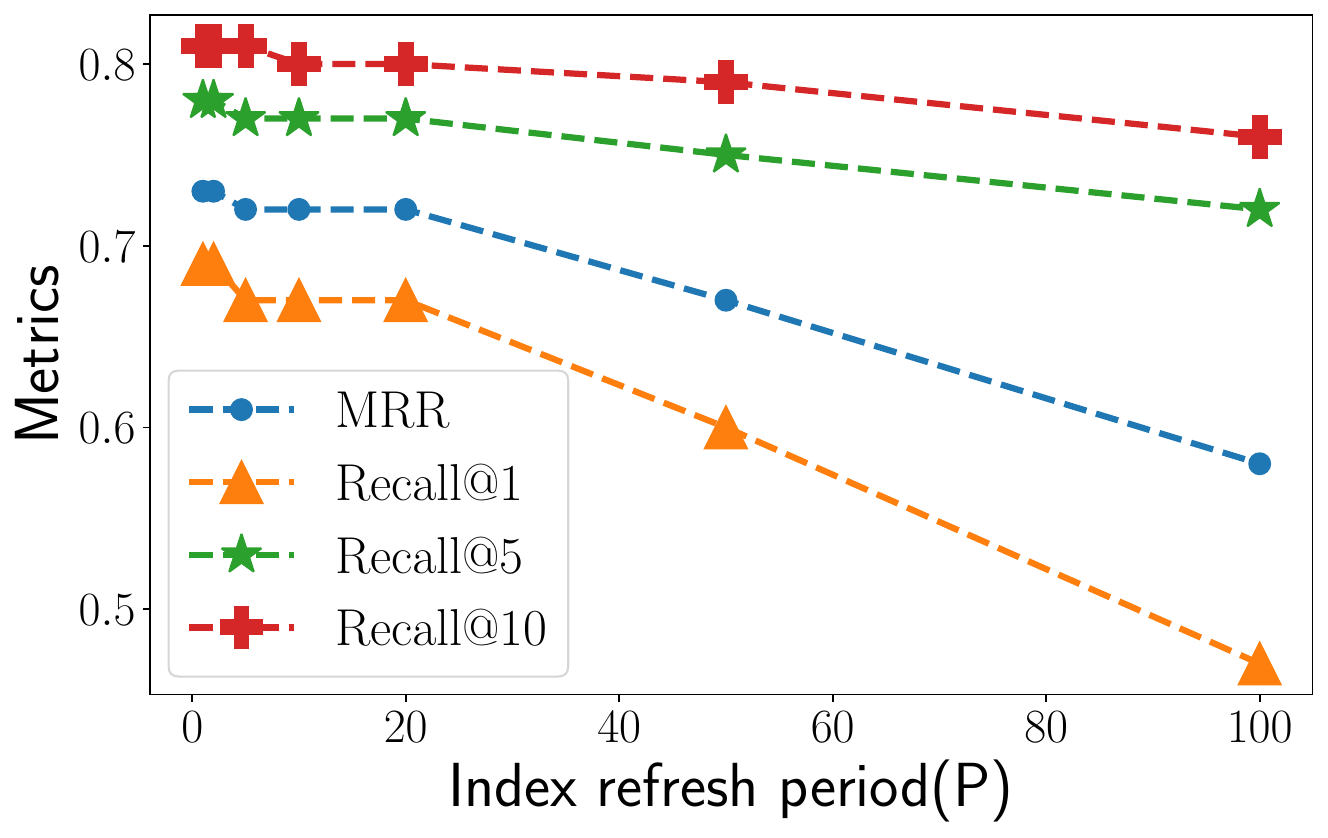}
\caption{Performance of proposed method when increasing the index frequency refresh period(P)}
\label{fig:ervarywiki}
\end{figure}
\subsection{Datasets}
\textbf{Wikipedia Edits:} It is a bipartite interaction graph between Wikipedia editors and pages. Each interaction is a timestamped edge containing \textit{LIWC} features of edited text \cite{jodie}.

\noindent \textbf{Reddit Posts:} Similar to \textit{wiki-edits}, it is a interaction graph between users and subreddits. Each interaction corresponds to a post written by a user in the corresponding subreddit. Each interaction is attributed with \textit{LIWC} features of post text \cite{jodie}. 

\noindent \textbf{Twitter Retweets:} We use the dataset released as part of rec-sys challenge \cite{twitter_challenge}. We keep only interaction of \textit{retweet} type and have interaction timestamp available. We keep the top 2000 creators with the most retweets and filter out those engagers with less than ten engagements. Interaction features are sentence embeddings of tweets computed using Bert \cite{bert}.

\noindent Table \ref{tab:dataset_stats} summarizes the dataset statistics.

\begin{table}[t!]
\caption{Dataset statistics}
\centering
\small
\begin{tabular}{lcccc}
\toprule
\textbf{Dataset} & \textbf{\# Nodes} & \textbf{\# Interactions} & \textbf{\# Features} & \textbf{\# Period} \\
\midrule
Wikipedia & $9227$ & $157474$ & $172$ & $720$ hrs. \\ 
Reddit & $11000$ & $672447$ & $172$ & $720$ hrs.\\  
Twitter & $4148$ & $36811$ & $768$ & $1$ week \\
\bottomrule
\end{tabular}
\label{tab:dataset_stats}
\end{table}

\subsection{Baselines - Temporal Graph Neural Networks}
There are many existing methods \cite{dysat,dyrep,jodie,tgn,tgat,pint} for modeling  on dynamic graphs. We integrate the proposed training procedure with \tgnns, \tgn\cite{tgn} and \tgat \cite{tgat}. These \tgnns learn the time-dependent representations for nodes that are utilized as an embedding index in our method. A recently proposed method \pint \cite{pint} uses positional features from the joint computational tree of source and target node to calculate the link formation probability. This requires calculating the features for every possible source-target node pair, making \pint ineffective in recommendation tasks. Moreover, our proposed method is not integrable with \pint due to the absence of target-independent node embeddings. For completeness, we report the performance of \pint in our experiments.


\noindent \textbf{Temporal Graph Attention Networks(\tgat) \cite{tgat}} computes the embedding of node $v$ at time $t$ by aggregating the features of temporal neighbourhood using attention. This denotes the one layer of message passing. Successive application of such layers constitutes multi-layer \textsc{Tgnn}. For more details, we refer to \cite{tgat}.

\noindent \textbf{Temporal Graph Networks(\tgn) \cite{tgn}} additionally stores the memory of each node apart from input features. These states are updated when the corresponding node is involved in an interaction using a recurrent neural network, e.g., LSTM/GRU \cite{lstm,gru}.

For both \tgnns, we use the default parameters as provided in \tgn code-base \footnote{https://github.com/twitter-research/tgn}. Moreover, we use the same chronological data split for training as default in \tgn codebase.

\begin{figure}[t!]
\centering
\includegraphics[width=.33\textwidth]{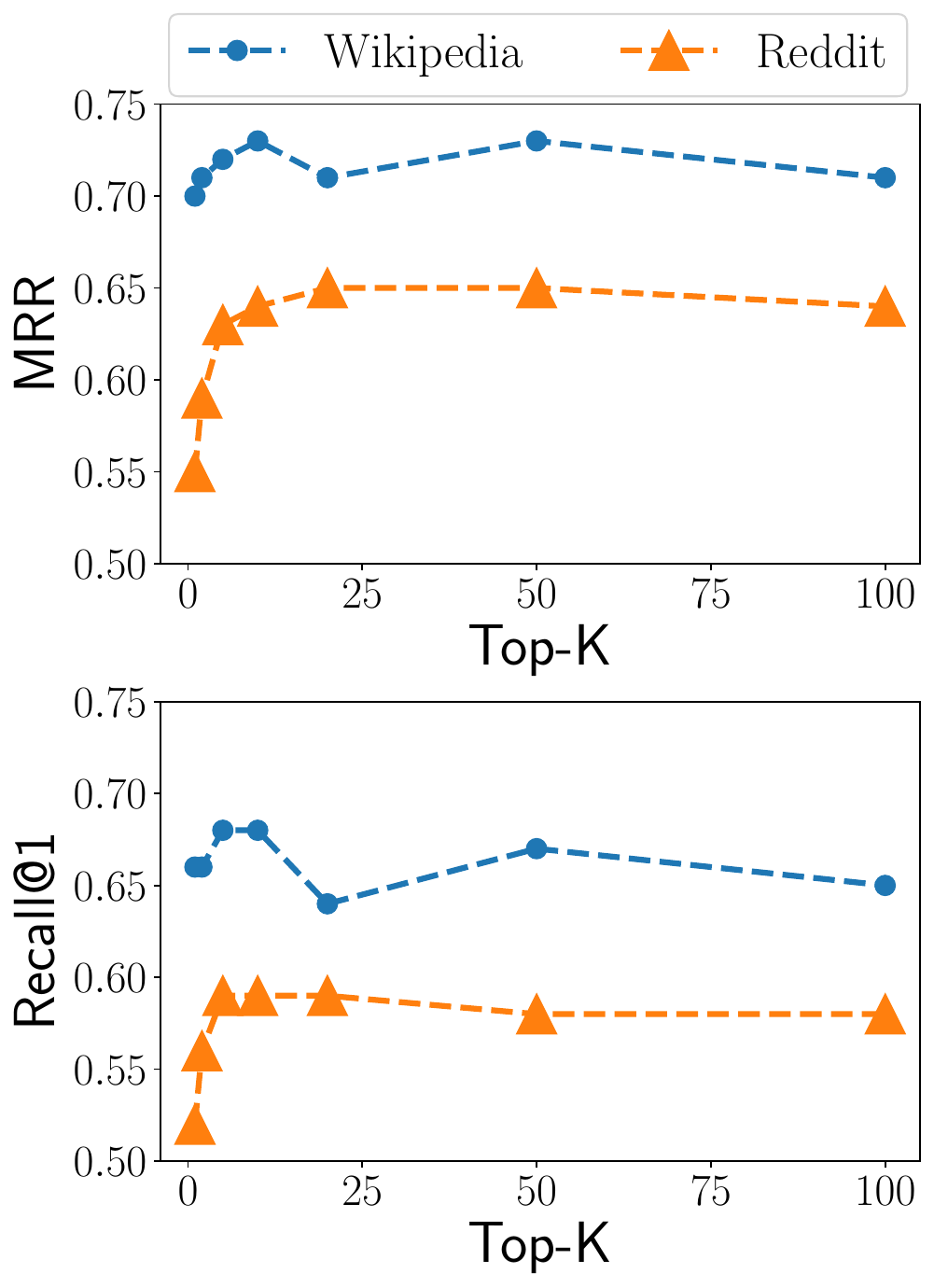}
\caption{Performance of proposed method when increasing the top-k in negative sampling}
\label{fig:vary_topk}
\end{figure}

\begin{figure}[b!]
\centering
\includegraphics[width=.40\textwidth]{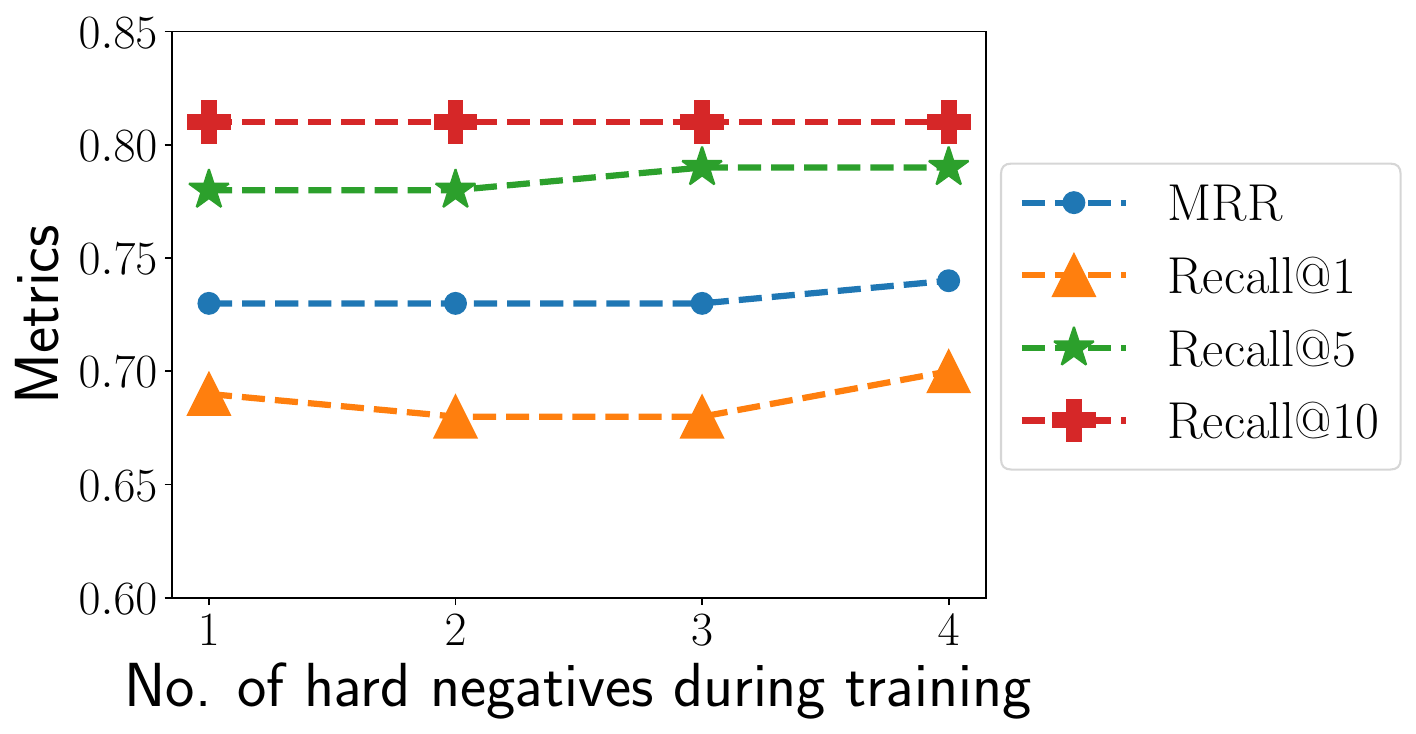}
\caption{Performance variation when varying the \# of hard negative samples during training \textsc{tgn} on Wikipedia dataset}
\label{fig:vary_hn}
\end{figure}

\subsection{Baselines - Sampling negative examples for \tgnns training} \label{sec:heuristichn}
We also designed two heuristic-based methods to sample negative nodes during the model training. Specifically, these method represents the two most known biases \cite{Zignani_Gaito_Rossi_Zhao_Zheng_Zhao_2014} regarding future link prediction. 

\noindent \textbf{Bias 1:} Two nodes with common neighborhoods have higher chances of connecting in the future.

\noindent \textbf{Bias 2:} Two nodes frequently interacting in the past have high chances of interacting again.

\begin{table*}[t!]
\caption{Performance comparison of various \tgnns based approaches. We report mean and std. dev. of 5 runs.}
\centering
\resizebox{0.7\textwidth}{!}{%
\begin{tabular}{llcccc}


\toprule
\textbf{Dataset} & \textbf{Method} & \textbf{MRR} &\textbf{Recall}\boldmath{$@1$} &  \textbf{Recall}\boldmath{$@5$} & \textbf{Recall}\boldmath{$@10$}\\
\midrule
\multirow{10}{*}{Wikipedia} & \tgat &$0.569 \pm 0.0097$& $0.5111 \pm 0.0135$& $0.6337 \pm 0.0056$& $0.6713 \pm 0.0058$\\ 
 & \tgn & $0.6154\pm0.0857$&$0.4934\pm0.1586$&$0.7512\pm0.0079$&$0.7963\pm0.007$\\
 & \pint & $0.7146 \pm 0.0043$ & $0.6521 \pm 0.0061$ & $0.7879 \pm 0.0031$ & $0.8189 \pm 0.0032$\\
 \cmidrule{2-6}
& \tgn+ Static HN & $0.5932 \pm 0.0596$& $0.5128 \pm 0.0935$& $0.6844 \pm 0.0193$& $0.7285 \pm 0.009$ \\
& \tgn+ IFQ HN & $0.4809 \pm 0.074$& $0.3646 \pm 0.1356$& $0.6116 \pm 0.0124$& $0.6576 \pm 0.0129$
 \\
  \cmidrule{2-6}
 &\tgat+ HN & $0.6132 \pm 0.0038$& $0.5844 \pm 0.0051$& $0.6417 \pm 0.0025$& $0.6623 \pm 0.0021$ \\ 
 & \tgn+ Nearest HN & $0.7032 \pm 0.0086$& $0.656 \pm 0.0116$& $0.7571 \pm 0.007$& $0.786 \pm 0.0071$\\
 &\tgn+ HN &  $0.7249 \pm 0.0038$& $0.6807 \pm 0.0057$& $0.7752 \pm 0.002$& $0.803 \pm 0.0011$ \\

& \tgn+ UN+HN  & \boldmath{$0.7339 \pm 0.0138$} & \boldmath{$0.6817 \pm 0.0254$}& \boldmath{$0.792 \pm 0.003$}& \boldmath{$0.8198 \pm 0.003$}\\
& \tgn+ UN+HN (F=10) & $0.7218 \pm 0.015$& $0.6709 \pm 0.0252$& $0.7781 \pm 0.0065$& $0.8075 \pm 0.006$\\
\cmidrule{1-6}
\multirow{10}{*}{Twitter} & \tgat & $0.1802 \pm 0.0051$& $0.1099 \pm 0.0045$& $0.2306 \pm 0.0078$& $0.3172 \pm 0.0061$ \\

& \tgn &$0.2928 \pm 0.006$& $0.1885 \pm 0.0053$& $0.3916 \pm 0.0104$& $0.5063 \pm 0.0086$
\\ 
& \pint & $0.4148 \pm 0.0041$& $0.297 \pm 0.0041$& $0.5519 \pm 0.0052$& $0.6497 \pm 0.0056$\\
\cmidrule{2-6}
& \tgn + Static HN & $0.1355 \pm 0.0138$& $0.0837 \pm 0.0121$& $0.1784 \pm 0.0172$& $0.2325 \pm 0.0153$\\ 

& \tgn + IFQ HN & $0.0754 \pm 0.0034$& $0.0506 \pm 0.0026$& $0.0958 \pm 0.0047$& $0.1162 \pm 0.0045$ \\

\cmidrule{2-6}

& \tgat + HN & $0.2061 \pm 0.0041$& $0.1346 \pm 0.0046$& $0.2686 \pm 0.0039$& $0.3515 \pm 0.0049$ \\
& \tgn + Nearest HN &$0.0976 \pm 0.0158$& $0.0486 \pm 0.0186$& $0.1152 \pm 0.0192$& $0.1681 \pm 0.0277$\\
 &\tgn+ HN &  $0.3192 \pm 0.0073$& $0.2229 \pm 0.0075$& $0.4163 \pm 0.0098$& $0.511 \pm 0.0092$\\
 & \tgn+UN+HN  & \boldmath{$0.3422 \pm 0.0081$}& \boldmath{$0.2431 \pm 0.0092$}& \boldmath{$0.4424 \pm 0.0084$}& \boldmath{$0.543 \pm 0.007$}\\
& \tgn+UN+HN (F=10)  & $0.3453 \pm 0.0068$& $0.2401 \pm 0.0068$& $0.4538 \pm 0.0077$& $0.5582 \pm 0.009$\\
\cmidrule{1-6}

\multirow{8}{*}{Reddit} & \tgat & $0.4251 \pm 0.0012$& $0.335 \pm 0.0015$& $0.5212 \pm 0.001$& $0.6034 \pm 0.0009$ \\

& \tgn &$0.6003 \pm 0.018$& $0.5204 \pm 0.0217$& $0.6889 \pm 0.0148$& $0.7463 \pm 0.0137$\\ 
&\pint & $0.5993 \pm 0.0027$& $0.5089 \pm 0.0032$& $0.7023 \pm 0.0022$& $0.7675 \pm 0.0014$\\

\cmidrule{2-6}
& \tgn + Static HN & $0.3561 \pm 0.0488$& $0.2448 \pm 0.0622$& $0.4813 \pm 0.0443$& $0.5578 \pm 0.0341$\\ 

& \tgn + IFQ HN & $0.4632 \pm 0.0245$& $0.4205 \pm 0.021$& $0.5052 \pm 0.0296$& $0.5365 \pm 0.0356$ \\

\cmidrule{2-6}

& \tgat + HN & $0.4631 \pm 0.0008$& $0.38 \pm 0.0014$& $0.5542 \pm 0.0009$& $0.6258 \pm 0.0013$ \\
& \tgn + Nearest HN &$0.552 \pm 0.0121$& $0.5178 \pm 0.0127$& $0.5837 \pm 0.013$& $0.6117 \pm 0.0131$\\
 &\tgn+ HN &  \boldmath{$0.6382 \pm 0.0124$} & $0.5749 \pm 0.0134$& \boldmath{$0.7074 \pm 0.0126$}& \boldmath{$0.7519 \pm 0.0125$}\\
 & \tgn+ UN+HN  & $0.6379 \pm 0.0135$& \boldmath{$0.5751 \pm 0.0133$}& $0.7059 \pm 0.0152$& $0.7518 \pm 0.0137$\\
 & \tgn+ UN+HN (F=10) & $0.6296 \pm 0.0182$& $0.5688 \pm 0.0178$& $0.6936 \pm 0.0195$& $0.7405 \pm 0.0194$
\\

\bottomrule
\end{tabular}%

}

\label{tab:main_results}

\end{table*}

Therefore, given an interaction $(u,v,t)$, potential candidates for negative node $v^{-}$ conform to bias 1 or 2 from source node $u$. And such a node should not be equal to node $v$. We next define the methods for computing negative nodes $v^{-}$ basis bias 1 and 2. 

\noindent\textbf{Static embedding based nearest samples:} We first convert training temporal graph $\mathcal{G}$ into a static graph $\mathcal{G}^{static} =(\mathcal{V}^{static},\mathcal{E}^{static})$ as follows.
\begin{gather}
\nonumber\mathcal{E}^{static} = \{(u,v) \mid \exists \; (u,v,t) \in \mathcal{G}_{train}\} \\ \mathcal{V}^{static} = \{u \mid \exists\; (u,v) \in \mathcal{E}^{static} \}
\end{gather}
We apply $2$-layer Graph Attention Network(\gat) \cite{gat} over $\mathcal{G}^{static}$ to compute node embeddings $\ch_v$ of all nodes $v \in \mathcal{V}^{static}$ using unsupervised link-prediction loss. The optimized embeddings using link prediction loss lead to community-based characteristics in node embeddings, i.e., nodes in the same community will be near in embedding space\cite{graphsage}. This captures the intuition of bias 1. Subsequently, we define the following probability distribution for sampling negative distribution for all source nodes:
\begin{equation}
    p_{{-}}^{static}(v) = \left\{ \frac{\exp(\text{cos}(\ch_v,\ch_u))}{\sum_{i \in V^{static}}\exp(\text{cos}(\ch_v,\ch_i))} \right\} \;\; \forall u \in V^{static}
    \label{eq:bias1}
\end{equation}
where $\ch_u \;\forall u \in V^{static}$ are learnt using $2$-layer \textsc{GAT}. During training \tgnns and given an interaction $(u,v,t)$, we sample the negative node $v^{-}$ from $p_{{-}}^{static}(v)$. In our experiments, we term this negative node sampling baseline as \textbf{Static HN}.

\noindent \textbf{Recent interactions based hard negatives:}
Given the temporal graph $\mathcal{G}$, we compute the frequency of nodes interacted before time $t$ by all source nodes $u$ in every interaction $(u,v,t) \in \mathcal{G}$, similarly to eq. \ref{eq:bias1}, we define the distribution over this frequency count and sample from this.  We term this baseline method as \underline{I}nteraction \underline{F}re\underline{q}uencies based \underline{H}ard \underline{N}egatives (\textbf{IFQ HN}).

\subsection{Results \& Discussions} 
We use recommendation metrics  \textbf{MRR}, \textbf{Recall@1}, \textbf{Recall@5}, and \textbf{Recall10} to evaluate \tgnns. Table \ref{tab:main_results} presents the performance of all methods across metrics in the link prediction task. To compute these metrics for every test interaction $(u,v,t)$, we rank all the graph nodes for the source node $u$ at time $t$ and compute the rank of the actual target node $v$ in the ranked result. This is repeated for all the test interactions, and metrics MRR, Recall@1, Recall@5, and Recall@10 are aggregated. In table \ref{tab:main_results}, methods \tgat and \tgn are \tgnns trained using the standard training procedure, which utilizes the negative examples sampled from the uniform sampling distribution. \tgn+ Static HN and \tgn+ IFQ HN methods are \tgn trained using heuristic-based hard negatives as described in section \ref{sec:heuristichn}. \tgat+ HN and \tgn + HN are \tgnns trained using the proposed negative sampling distribution. \tgn+ Nearest HN method is trained by assigning the nearest node (apart from target node $v$) to source node $u$ as the hard negative example during training. We use this method to show the importance of computing hard negatives from top-$K$ nearest nodes. Finally, we introduce a hybrid training procedure that utilizes $2$ negative examples, one from \underline{U}niform distribution, which is standard practice, and one from the proposed distribution. We term this as \tgn+UN+HN.

\begin{figure}[t!]
\centering
\includegraphics[width=.35\textwidth]{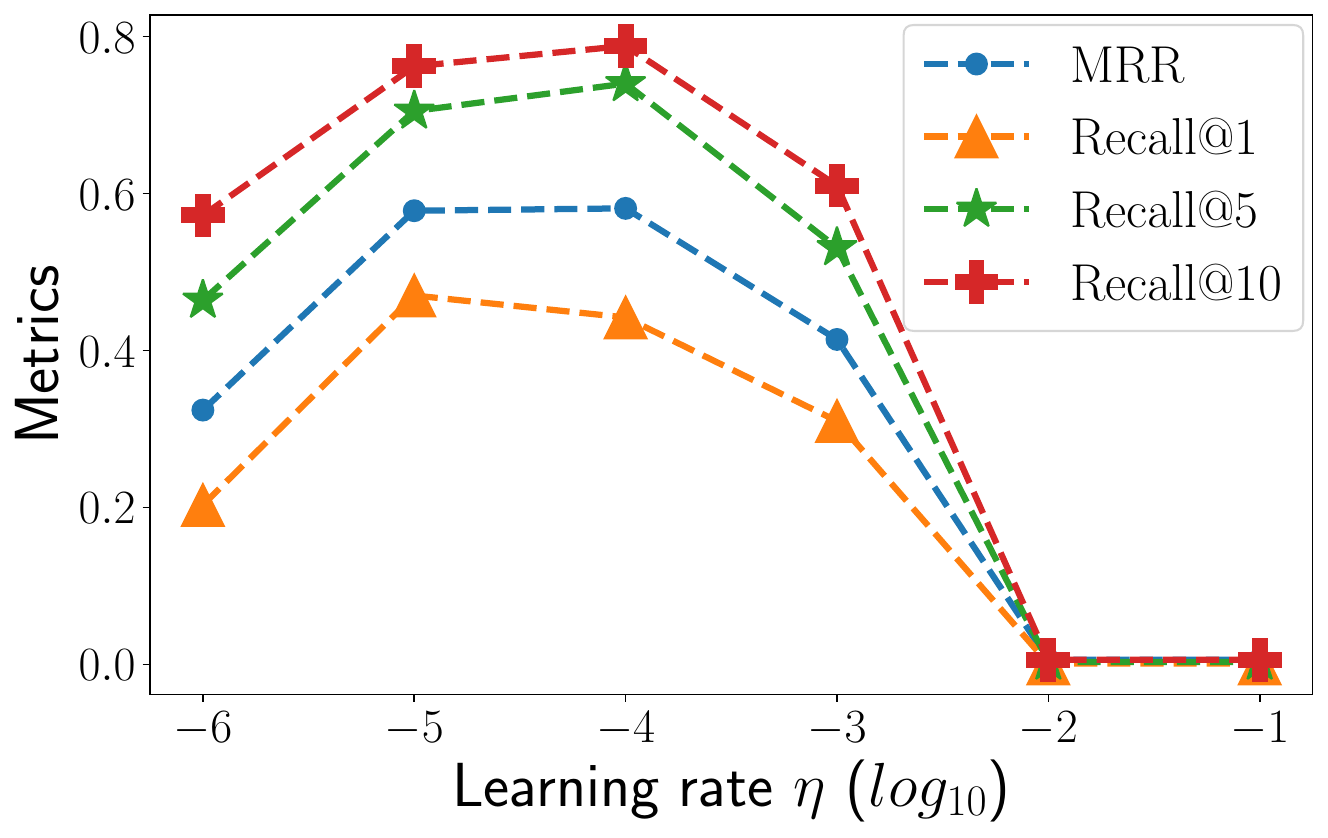}
\caption{Influence of varying learning rate on \tgn performance on Wikipedia dataset}
\label{fig:lrrnwiki}
\end{figure}

\begin{figure}[t!]
\centering
\includegraphics[width=.35\textwidth]{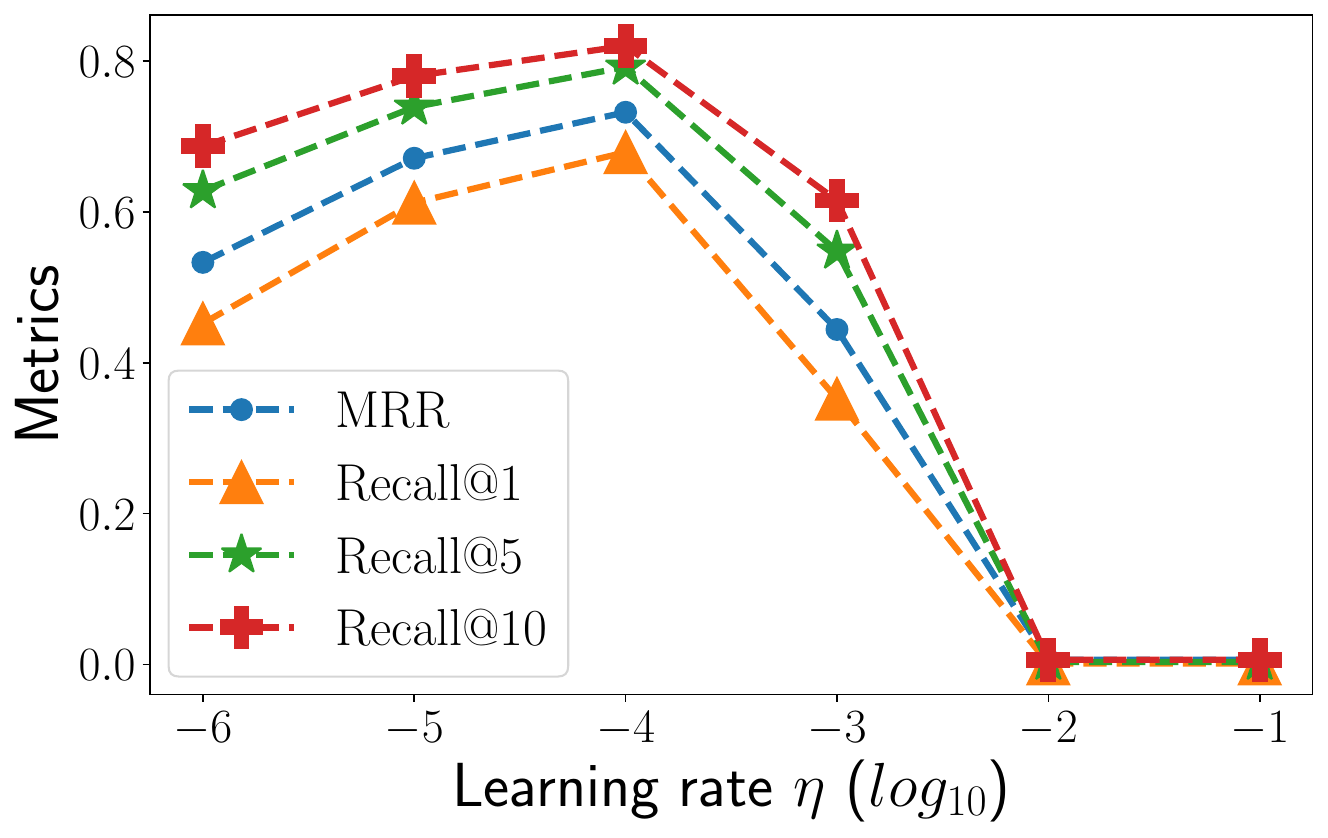}
\caption{Influence of varying learning rate on proposed model training procedure (\tgn+UN+HN) on Wikipedia dataset}
\label{fig:lrhnwiki}
\end{figure}

Table \ref{tab:main_results} clearly shows that the proposed training procedure produces consistent gains over the standard \tgnns training methods across all datasets and metrics.  \tgn+ HN and \tgat+ HN provides better performance than \tgn and \tgat respectively. Moreover, \tgn+ Nearest HN performs worse than \tgn+ HN showcasing the importance of sampling from top-$k$ nearest nodes from the source node while sampling the hard negatives instead of using the nearest one. We also see that heuristic-based negative examples mining baselines \tgn+ Static HN and \tgn+ IFQ HN perform much worse than even \tgn trained using sampling from the uniform distribution. This shows that heuristic-based methods are sub-optimal in training the \tgnns. Finally, the hybrid training procedure \tgn+ UN + HN, which utilizes negative samples from both the uniform distribution and the proposed distribution, performs marginally better than \tgn + HN, especially in the Twitter dataset, indicating the importance of such hybrid methods.

\begin{figure}[t!]
\centering
\includegraphics[width=.35\textwidth]{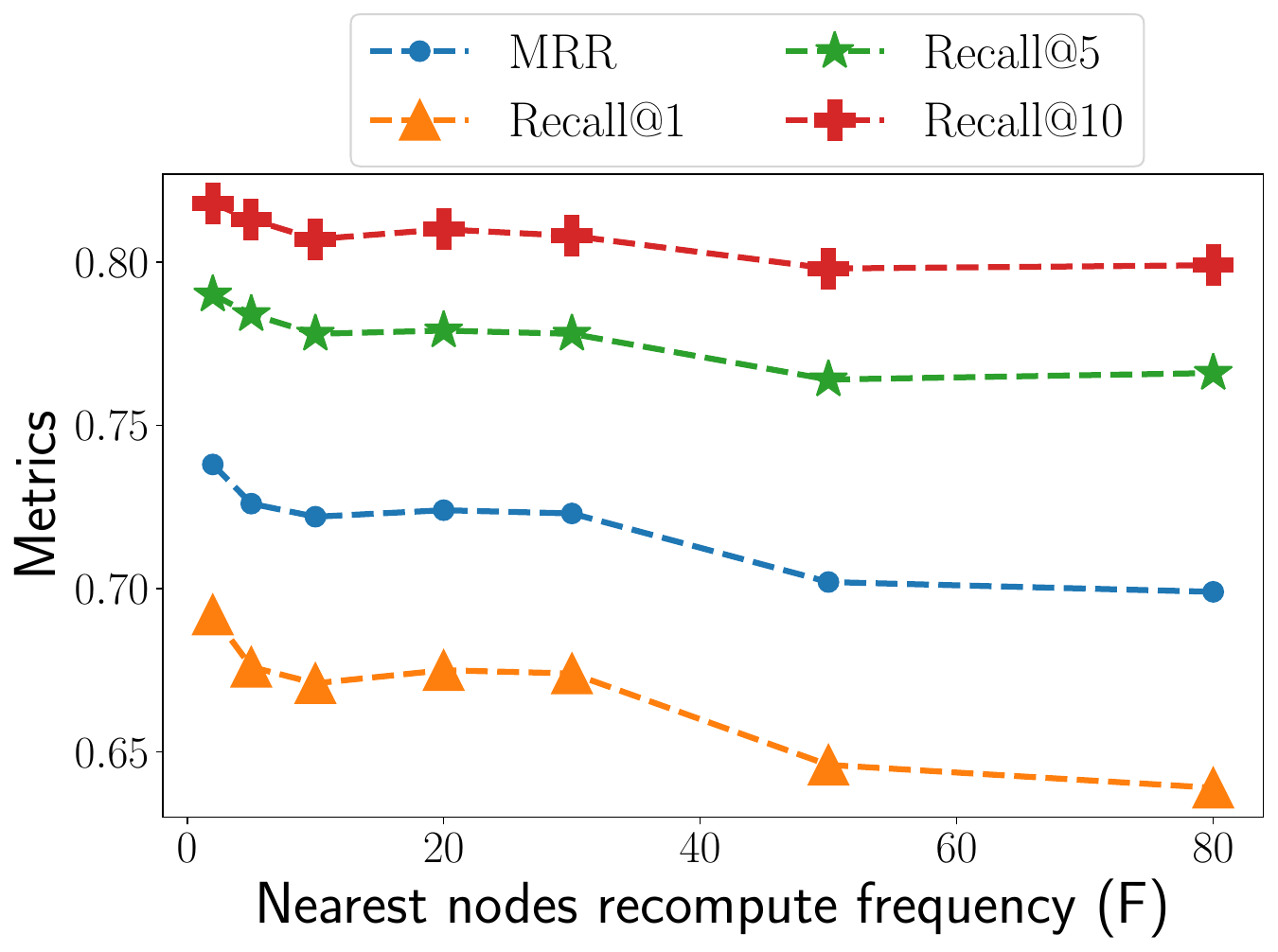}
\caption{Impact on model performance on varying $F$ on Wikipedia dataset}
\label{fig:varyFwiki}
\end{figure}

Additionally, we vary the index refresh period $P$, which recomputes the node embedding after every $batch\_id \;\%\;P$ during training. As seen in figure \ref{fig:ervarywiki}, performance remains marginally stable at the low refresh period till $20$ in the Wikipedia dataset. If $P$ increases to $50$ or $100$, performance is decreased drastically. We also vary the top-$k$ in figure \ref{fig:vary_topk} and find that while in Wikipedia data, top-k=$5$ provides the best results, it revises to $20$ in the Reddit dataset. Finally, we found that sampling more than one hard negative example for each interaction during training doesn't benefit the model performance. Figure \ref{fig:vary_hn} clearly shows that varying the \# of hard negatives doesn't significantly impact the metrics on test data. $\eta$ is an important hyper-parameter, as seen in eq. \ref{eq:parameter_convergence}. Thus, we vary $\eta$ in \tgn and \tgn+UN+HN to analyze its impact on model performance. In figures \ref{fig:lrrnwiki} and \ref{fig:lrhnwiki}, we observe that very high and very low $\eta$ results in worse performance in both standard model training and proposed model training. In table \ref{tab:training_time}, we furthermore show the training time of \tgn, \pint, proposed method \tgn+UN+HN, and its variants where we only re-calculate $\text{top-K}$ nearest nodes at epoch frequency $F=5$,$10$ and $20$. Though the proposed method significantly increases training time, its proposed variants are comparatively faster with a lower accuracy drop, as seen in figure \ref{fig:varyFwiki} on the wiki dataset.

We also emphasize that the increased training time is irrelevant during inference as our proposed training approach doesn't modify the inference process and thus, inference time.

\begin{table}[t!]
\caption{Training time analysis(hours)}
\centering
\small
\begin{tabular}{lccc}
\toprule
\textbf{Method} & \textbf{Wikipedia} & \textbf{Reddit} & \textbf{Twitter}  \\
\midrule
\tgn & $0.36$ & $1.74$ & $0.29$  \\ 
\pint & $10.62$& $103.67$ & $0.82$ \\
\tgn+UN+HN & $2.17$ & $11$ & $1.20$ \\ 
\tgn+UN+HN (F=5) & $0.85$ & $4.58$ & $0.62$ \\
\tgn+UN+HN (F=10) & $0.66$ & $4.23$ & $0.55$  \\
\tgn+UN+HN (F=20) & $0.59$ &  $3.00$ &  $0.51$\\
\bottomrule
\end{tabular}
\label{tab:training_time}
\end{table}

\section{Conclusion}
In this work, we highlighted that the current training technique of \tgnns is sub-optimal, resulting in unsatisfactory performance on end tasks, especially in recommendations-based applications where the precise ranking of target nodes is critical. To remedy this, we analyzed the effect of negative nodes required in the loss calculation with parameter convergence. Subsequently, we proposed a dynamic probability distribution to sample the negative nodes for learning optimal parameters. Through extensive evaluation of 3 real-world datasets, we established that our proposed training procedure results in better performance across relevant metrics over the test temporal graphs.

\bibliographystyle{ACM-Reference-Format}
\bibliography{references}

\end{document}